\documentclass[conference]{IEEEtran}
\IEEEoverridecommandlockouts
\usepackage{amsmath,amssymb,amsfonts}
\usepackage{algorithmic}
\usepackage{graphicx}
\usepackage{textcomp}
\usepackage{xcolor}
\usepackage{booktabs}
\usepackage[style=ieee]{biblatex} 
\def\BibTeX{{\rm B\kern-.05em{\sc i\kern-.025em b}\kern-.08em
    T\kern-.1667em\lower.7ex\hbox{E}\kern-.125emX}}
\begin{document}

\title{L-WaveBlock: A Novel Feature Extractor Leveraging Wavelets for Generative Adversarial Networks }

\author{\IEEEauthorblockN{Mirat Shah*, Vansh Jain*, Anmol Chokshi*, Guruprasad Parasnis*, Pramod Bide*}

\IEEEauthorblockA{Sardar Patel Institute of Technology, All authors contributed equally*}}

\maketitle

\begin{abstract}
Generative Adversarial Networks (GANs) have risen to prominence in the field of deep learning, facilitating the generation of realistic data from random noise. The effectiveness of GANs often depends on the quality of feature extraction, a critical aspect of their architecture. This paper introduces L-WaveBlock, a novel and robust feature extractor that leverages the capabilities of the Discrete Wavelet Transform (DWT) with deep learning methodologies. L-WaveBlock is catered to quicken the convergence of GAN generators while simultaneously enhancing their performance. The paper demonstrates the remarkable utility of L-WaveBlock across three datasets, a road satellite imagery dataset, the CelebA dataset and the GoPro dataset, showcasing its ability to ease feature extraction and make it more efficient. By utilizing DWT, L-WaveBlock efficiently captures the intricate details of both structural and textural details, and further partitions feature maps into orthogonal subbands across multiple scales while preserving essential information at the same time. This streamlined process translates into accelerated results, elevating the overall efficiency of GAN-based image generation. Not only does it lead to faster convergence, but also gives competent results on every dataset by employing the L-WaveBlock. The proposed method achieves an Inception Score of 3.6959 and a Structural Similarity Index of 0.4261 on the maps dataset, a Peak Signal-to-Noise Ratio of 29.05 and a Structural Similarity Index of 0.874 on the CelebA dataset. The proposed method performs competently to the state-of-the-art for the image denoising dataset, albeit not better, but still leads to faster convergence than conventional methods. With this, L-WaveBlock emerges as a robust and efficient tool for enhancing GAN-based image generation, demonstrating superior convergence speed and competitive performance across multiple datasets for image resolution, image generation and image denoising.

\end{abstract}


\section{Introduction}

In recent years, Generative Adversarial Networks (GANs) have developed into a strong base for a range of image creation and image enhancement jobs. In a variety of computer vision-related domains, their ability to comprehend complicated data distributions and generate high-quality images has encouraged development. Enhancing feature extraction, quickening convergence, and ensuring consistent performance across varied image-altering jobs are ongoing challenges. Fig 1 shows the general architecture of the GAN.

In this study, we propose a novel approach that, by utilizing the capabilities of the DWT, revolutionizes feature extraction in GANs. Though DWT's typical uses are in signal processing and picture compression, we employ its multi-resolution analytical capabilities to enhance feature extraction in the context of GANs. 
The formula for the Daubechies 2 (db2) DWT in 1D is as follows: 

Let $x[n]$ be the input signal and $x'[n]$ be the approximation (LL) at a certain level $j$ and $y[n]$ be the detail (LH, HL, or HH) at that level:

\[
x'[n] = \sum_{k} h[k] \cdot x[2n - k]
\]

\[
y[n] = \sum_{k} g[k] \cdot x[2n - k]
\]

In these equations:

- $j$ represents the scale or level of the DWT.
- $k$ represents the filter tap index.
- $h[k]$ and $g[k]$ are the filter coefficients for the low-pass (scaling) and high-pass (wavelet) filters, respectively, associated with the db2 wavelet.

The db2 wavelet has four filter coefficients, which are typically defined as follows:

\[
h[k] = \frac{1}{2} \cdot (1 + \sqrt{3}) \text{ for } k = 0, 1
\]
\[
h[k] = \frac{1}{2} \cdot (3 + \sqrt{3}) \text{ for } k = 2, 3
\]
\[
g[k] = (-1)^k \cdot h[3-k] \text{ for } k = 0, 1, 2, 3
\]

These filter coefficients are used in the convolution operation to obtain the approximation and detail coefficients at each level of the db2 DWT. The DWT can be applied iteratively to further decompose the approximation component at each level, resulting in a multi-resolution analysis of the input signal.

Our methodology stands out for its remarkable flexibility and adaptability, driving improvements in three pivotal use cases:

1. Conversion of a satellite image to a map image:\par
The conversion of high-resolution satellite photos into comprehensive, contextually accurate maps is one of the main issues in satellite image analysis. Traditional GANs frequently struggle to capture fine-grained spatial features and encounter lengthy convergence times. Our DWT-enhanced GAN excels at maintaining complex spatial information while also accelerating convergence. By including DWT, the GAN becomes a feature extraction powerhouse that can interpret the subtleties of satellite images and provide highly accurate map representations. \par

\break
2. Image Denoising: \par
The fundamental principle of image processing has long been picture enhancement through noise reduction. Traditional methods, however, can obscure important details while also reducing noise. By adding DWT, our GAN design adds a novel feature that makes it successfully separate noise from the signal. Our GAN excels at extracting noise-free, contextually important properties by utilizing the multi-resolution analysis provided by DWT, ushering in a revolutionary new age for denoising techniques. \par

3. Improving Resolution of Images: \par
Enhancing image resolution is essential for many uses, including surveillance and medical imaging. The difficult component is converting low-resolution photos into high-resolution images while maintaining essential structural and textural information. The goal of resolution improvement utilizes the feature extraction prowess of our DWT-enhanced GAN. The GAN skillfully upscales photos by smoothly incorporating DWT, producing results that go beyond simple pixel interpolation by accurately capturing the essence of the underlying data. \par

Our research highlights DWT's function in accelerating convergence and increasing feature extraction. We offer a thorough technique that makes use of DWT's multi-resolution analysis capabilities, enabling GANs to perform very well across a range of picture editing applications. This study highlights the flexibility of our method and its revolutionary potential in numerous computer vision fields. The strong effects of DWT-driven feature extraction on GAN performance are supported by our empirical assessments, which are supported by the convincing outcomes in the aforementioned applications.

\begin{figure}
\centering
  \includegraphics[width=1\linewidth]{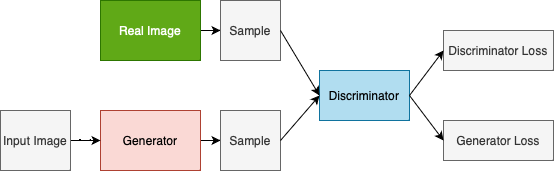}
  \caption{GAN Architecture}
  \label{fig1}
\end{figure}

\section{Related Work}
A lot of work has been undertaken in the field of GANs and how they are leveraged to attain state-of-the-art results in image denoising, image generation, etc. GAN was first introduced by Goodfellow et al. \cite{1} which started the revolution of this network. Tran et al. \cite{2} use GANs in the most basic way. The “clean” images are transformed into raw images through unprocessing. With the subsequent addition of noise, the resulting images are used to train a GAN to produce a noise model. A deep network denoiser uses the GAN output (noise) and clean images to learn how to map (create) clean images from noisy images. Similar efforts to effect the denoising of images were carried out by Wang et al. \cite{3}. They propose a DRGAN model presented for image denoising, which achieves good perceptual quality
with respect to both overall content and specific details. The trained model performs amazing effects not only on single image denoising tasks but also on other aspects such as defogging. Their experiments show that the model can also process grayscale images, presenting stronger texture and brightness.
The idea of using neural networks for denoising tasks was not constrained to GANs as demonstrated by Burger et al. \cite{4} by their use of Multi-Layer perceptrons for image denoising applications. Tripathi et al. \cite{5} show that latent vector recovery from GAN, in practice, can be used for image denoising. Alsaiari et al. \cite{6} demonstrated a GAN-like structure for image denoising, which was closely similar to ResNet. A generator network was trained to generate noise-free images by going against a discriminator network, using ground truth reference images to improve the quality of the generated images. They make use of residual blocks, skip connections, and batch normalization. Having a larger number of residual blocks increased the training accuracy significantly but with the incurred expense of requiring longer training times. Apart from image denoising, there have been concerted efforts to leverage techniques like wavelet transform for efficient feature extraction and also to reduce model complexities. Ilyas et al. \cite{7} make use of the DWT technique for feature extraction from facial expressions and achieve commendable results. To reduce model complexity and gain comparable results to conventional CNNs with pooling layers, Liu et al. \cite{8} proposed a multi-level wavelet CNN, which can also be extensively used for image-denoising tasks.

Fu et al. \cite{10} propose the DW-GAN for image denoising using DWT techniques. They introduce the DWT block within their proposed architecture and evaluate the model on various benchmark datasets. 
A network that learns various loss functions and responds to multiple settings when employed in a GAN is showcased by Isola et al. \cite{11}. There have also been methods for extracting and processing textures effectively through GANs. A good example of this is Li et al. \cite{14}, who propose a Markovian Generative Adversarial Networks (MGANs), a method for training generative neural networks for efficient texture synthesis. GANs are also put to good use for effective training purposes as demonstrated in the various methods employed by Salimans et al. \cite{15}, using which they achieved state-of-the-art results in semi-supervised classification tasks on famous datasets. As demonstrated by Kodali et al. \cite{16}, the use of GANs can also be done to achieve fast convergence of models without the use of much complexity. Lastly, a detailed overview of the applications and uses of GANs in denoising, generation and efficiency tasks is given by Aggarwal et al. \cite{17}. 
\par
In the realm of Super-Resolution Generative Adversarial Networks (SRGANs), a major breakthrough was achieved with the work of Ledig et al. \cite{18}. Their SRGAN introduced the concept of utilizing adversarial training to generate photo-realistic high-resolution images from low-resolution inputs. SR-GAN was particularly useful due to the integration of perceptual loss, an ingenious amalgamation of conventional pixel-wise mean squared error and content loss derived from feature maps obtained through a pre-trained VGG network. This approach greatly improved the visual authenticity of super-resolved images, rendering them very realistic.

Building upon this foundation, Lim et al. \cite{19} presented Enhanced Deep Residual Networks (EDSR) for single image super-resolution. EDSR ingeniously blended residual blocks with deep CNNs, thereby improving the model's proficiency in capturing intricate image details. The outcomes were substantial, establishing EDSR as a pioneer in achieving state-of-the-art results, as measured by the Peak Signal-to-Noise Ratio (PSNR) and the Structural Similarity Index (SSIM).

Subsequently, Wang et al. \cite{20} increased the capabilities of SRGANs with the development of Enhanced Super-Resolution Generative Adversarial Networks (ESRGAN). Relativistic Average GAN (RaGAN) loss function, spectral normalization, and a perceptual loss based on a specific feature extractor were all developed by ESRGAN as part of a unique architecture. As a result of this architectural improvement, ESRGAN was able to produce high-resolution images with remarkable sharpness and aesthetic appeal. As a result, ESRGAN not only established itself as a standard in the industry but also highlighted the crucial function that perceptual loss functions play in Super-Resolution GANs. Together, these influential studies demonstrate important developments in picture super-resolution, demonstrating the effectiveness of adversarial training, architectural breakthroughs, and loss function design in producing world-class super-resolution outcomes.

\begin{figure*}
\centering
  \includegraphics[width=1\linewidth]{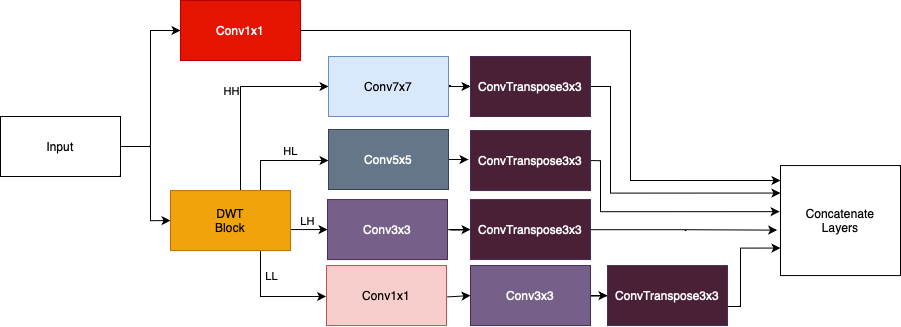}
  \caption{L-WaveBlock Architecture}
  \label{fig2}
\end{figure*}
\section{Proposed Methodology}


Our model is an adaptation of the generator-based UNet architecture that is frequently used in GANs. In this work, we develop a special skip connection method that makes use of the DWT to increase feature extraction and boost generator performance. We employed the Daubechies 2 wavelet in the DWT.
Fig 2 shows the proposed architecture of the L-WaveBlock.
When a skip connection is discovered in our model, it is regarded as the "input." This input is transformed using the DWT rather than being passed directly. The input is divided into four components: LL (Low-Low), LH (Low-High), HL (High-Low), and HH (High-High), each of which stands for a particular frequency band and piece of geographic data. The four components are then processed separately through two convolutional layers after the DWT. In order to extract pertinent features from the decomposed components, these layers are essential. A convolutional transpose layer is then applied to the outputs from the convolutional layers for each component. Up-sampling and spatial information restoration are both aided by this process. In parallel, the skip connection's original input also goes through processing with a different convolutional layer. By doing this, the input data is made sure to be seamlessly incorporated into the feature extraction procedure. The outputs from all five routes (LL, LH, HL, HH, and the processed input) are concatenated to produce an extensive feature representation. Then, a skip connection is made using this feature map that has been concatenated.


The reasons for proposing this methodology are as follows: 
\begin{enumerate}
    \item \textbf{Choice of using different numbers of convolutional layers for different bands of DWT output}: \par
    Due to the varying complexity of these components, it was decided to use two convolutional layers for the LL component and just one for the high-pass (LH, HL, and HH) components in the model. With additional layers and smoother, low-frequency input, LL effectively learns hierarchical representations. High-pass components, on the other hand, just need one layer to avoid over-processing and computational overhead, ensuring the retention of tiny features during feature extraction. They are rich in detailed high-frequency details. This strategy strikes the ideal balance between computational effectiveness and feature extraction.

    \item \textbf{How using DWT in the middle of skip connection helps}: \par Incorporating DWT in the middle of a skip connection within a UNet-based generator for a GAN offers multiple advantages that can be leveraged. DWTs possess the ability to decompose the skip connection input into multiple frequency components to enhance the feature representation. This enables the generator to capture both global and local details simultaneously, which is key to generating realistic images in a GAN. By applying DWT, the model learns multi-scale information, increasing its capacity to generate diverse and high-quality images. Moreover, this approach leads to efficient feature extraction and spatial detail conservation, leading to faster and more accurate image synthesis, a significant part of successful GAN training. Additionally, the integration of DWT within the skip connection aids convergence by providing the generator with detailed, multi-scale features, helping the GAN learn intricate image details with lesser time complexity, thereby reducing training time.

\end{enumerate}

Fig 3 shows the output for multiple subbands of an image in the CelebA dataset. 
\begin{figure}[h]
\centering
  \includegraphics[width=0.5\textwidth]{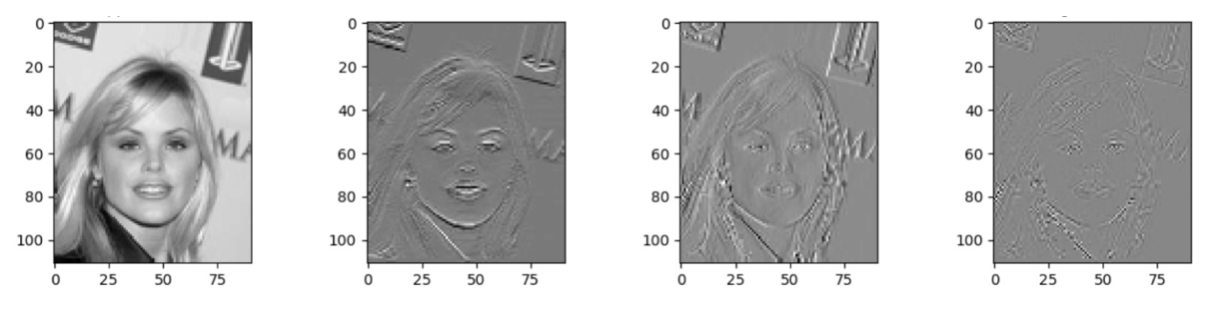}
  \caption{DWT output of CelebA Dataset(LL, LH, HL, HH)}
  \label{fig3}
\end{figure}

\section{Experimental Results}
\subsection{Dataset}

\begin{enumerate}
  \item \textbf{Maps Dataset}: The maps dataset contains paired pictures made for image-to-image translation tasks and is frequently used in computer vision research. Each pair consists of two images: a "map" image that represents a map or sketch of a particular location and a "satellite" image that shows the same location from a satellite's perspective. This dataset stands out for its diversity, which includes a range of natural landscapes, urban settings, bodies of water, and roadways. It is easier to couple data when map and satellite photos have the same $(512 \times 512)$ resolution. Both 1096 train pairs and 1096 test pairs are present. Fig 4 shows a sample image from the maps dataset.

  \begin{figure}[htbp]
\centering
\includegraphics[width=0.4\textwidth]{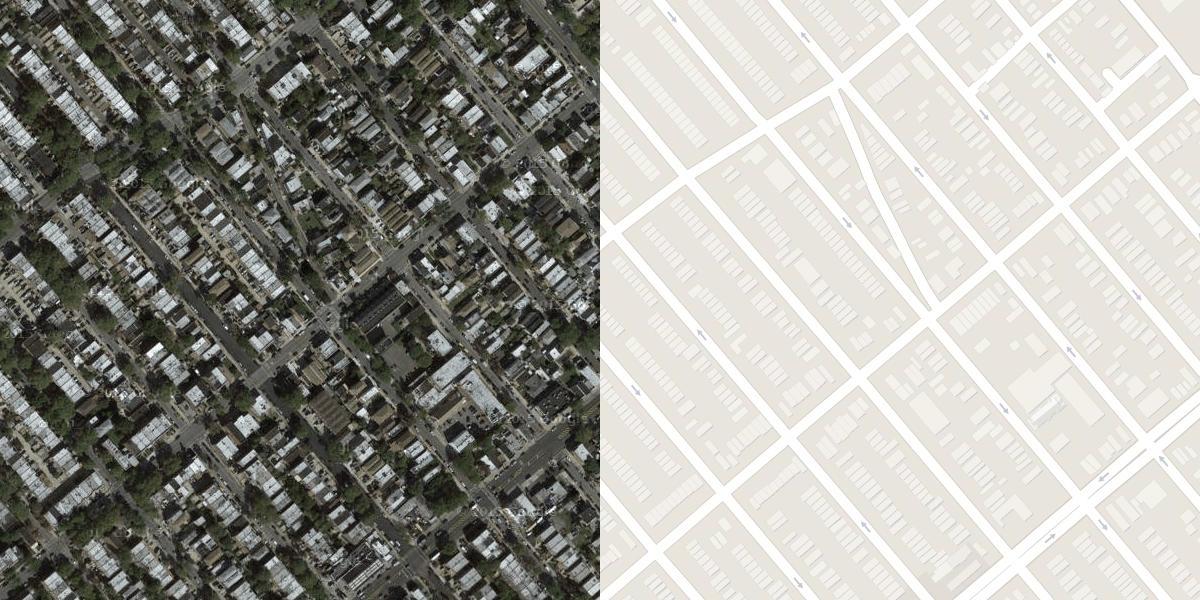}
\caption{Satellite image of the dataset}
\label{fig4}
\end{figure}

  \item \textbf{CelebFaces Attributes Dataset (CelebA)}: CelebA is a large-scale face attributes dataset with more than 200,000 celebrity images, each with 40 attribute annotations. This dataset of photos includes a wide range of poses and cluttered backgrounds. With 10,177 identities, 202,599 face photos, 5 landmark locations, and 40 binary attribute annotations per image, CelebA has a wide variety, a huge quantity, and rich annotations. Fig 5 shows a sample image from the CelebA dataset. 

 \begin{figure}[htbp]
    \centering
    \includegraphics[width=0.45\textwidth]{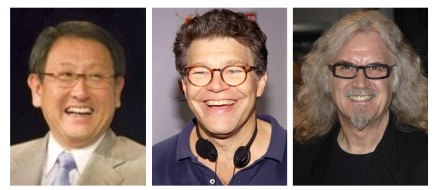}
    \caption{Sample image from the CelebA Dataset \cite{9}}
    \label{fig5}
    \end{figure}

  \item \textbf{GoPro dataset}: The GoPro dataset for deblurring consists of 3,214 blurred images with the size of $(1,280 \times 720)$ that are divided into 2,103 training images and 1,111 test images. The dataset consists of pairs of realistic blurry images and the corresponding ground truth shape images that are obtained by a high-speed camera. Fig 6 shows the sample image from the GoPro dataset. 

   \begin{figure}[htbp]
    \centering
    \includegraphics[width=0.45\textwidth]{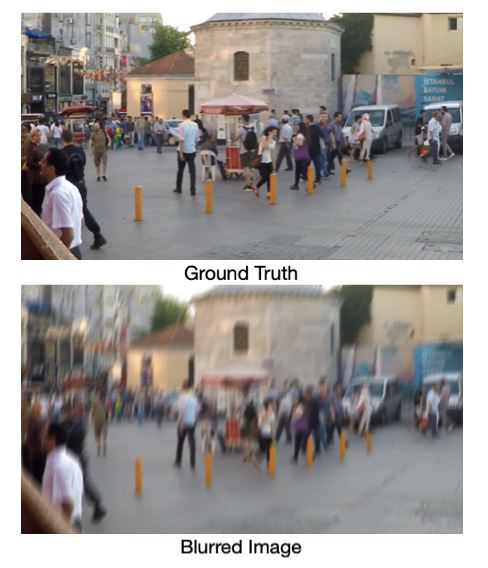}
    \caption{Sample image from the GoPro Dataset}
    \label{fig6}
    \end{figure}


\end{enumerate}








\subsection{Performance Metrics}
Pixel accuracy, Fréchet Inception Distance (FID), and Inception Score (IS) are performance indicators for generative AI tasks. The IS evaluates the value and variety of the samples that are produced. It assesses the generated images using an Inception classifier and determines a score that judges the samples' quality and diversity. Better performance is indicated by a higher value of IS. Another statistic that evaluates the standard and variety of generated images is FID. Using the Inception network, it compares how features are distributed in actual and artificial images. Better performance is indicated by a lower FID score. Based on pixel-level comparisons between generated and actual images, pixel accuracy measures the quality of an image. Almeida et al. \cite{13} additionally make other measurements for enhancing measures. 


\subsection{Results on the road imagery dataset}
Using a CNN with the DWT in the road imagery dataset offers several advantages over a normal CNN in generative tasks. DWT allows the decomposition of the road image into different frequency sub-bands, creating a multiscale representation of the image. This can capture both fine and coarse details in the image, making it easier to identify features at different scales.
It can help localize features in an image effectively. By decomposing the image into different frequency components, it was found that in the road imagery dataset, focusing on specific sub-bands that contain relevant information for a particular task, improves feature localization. 
It can be seen from Table I that when the generator is used without the proposed L-WaveBlock, the IS is found to be 3.6337 and the SSIM is 0.4108. In contrast, if the L-WaveBlock is employed with the generator, the IS is found to be 3.6959 and the SSIM obtained is 0.4261, which is slightly better than the results obtained without the proposed approach. Fig 7 illustrates how the suggested method produces an output that closely resembles the expected one on the road dataset. An Inception score of 3.6959 suggests that the generated images exhibit competent diversity and quality, with higher scores indicating better image quality. Not circumscribing the quantitative results, Fig 8 shows the loss convergence for the generator without and with the L-WaveBlock respectively for the road imagery satellite dataset. The sharp and faster convergence of the L-WaveBlock is clearly seen, demonstrating a superior efficiency over other feature extraction techniques.

\begin{figure}[htbp]
\centering
\includegraphics[width=0.5\textwidth]{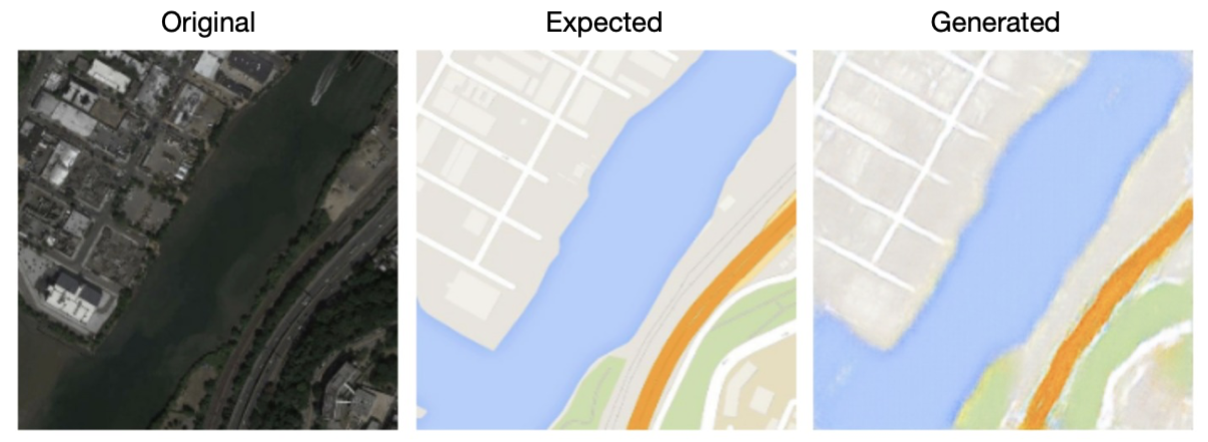}
\caption{Generated output for the road dataset}
\label{fig7}
\end{figure}

DWT also minimizes the amount of data that needs to be processed at each network tier. This results in a decrease in computational complexity, which is shown in this research to be very beneficial. As part of the decomposition procedure, the suggested approach additionally denoises the photos. A significant portion of the research is focused on denoising throughout the generating phase. 

\begin{figure}[htbp]
\centering
\includegraphics[width=0.5\textwidth]{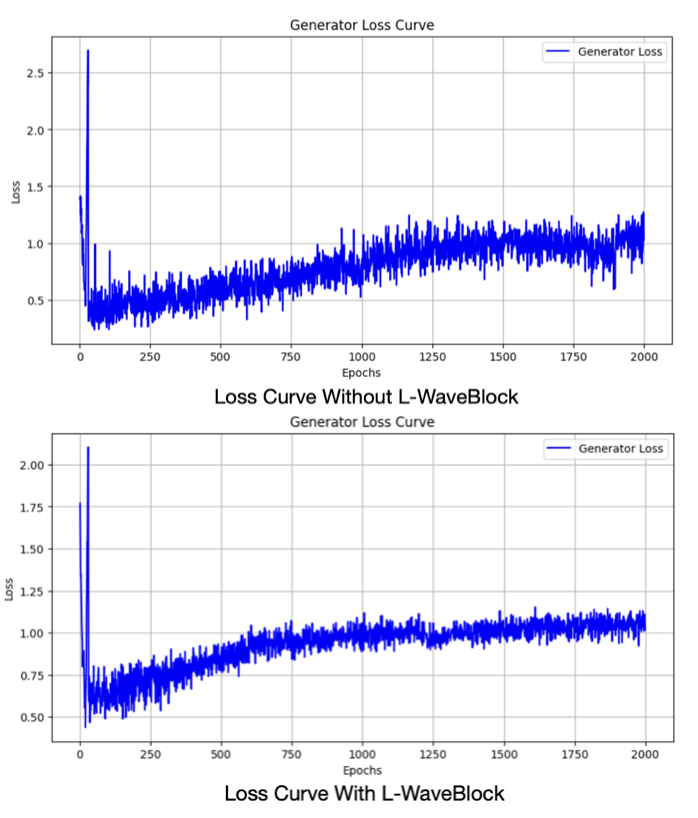}
\caption{Loss curve with and without L-WaveBlock for maps imagery dataset}
\label{fig8}
\end{figure}

\begin{table}
\centering
\caption{Performance metrics on the Maps dataset}
\begin{tabular}{lcc}
\toprule
\textbf{Architectures} & \textbf{IS} & \textbf{SSIM} \\
\midrule
\textbf{Generator without L-WaveBlock} & 3.6337 & 0.4108 \\
\textbf{Generator with L-WaveBlock} & \textbf{3.6959} & \textbf{0.4261} \\

\bottomrule
\end{tabular}
\end{table}

Additionally, DWT lessens the dimensionality of the input data, which aids in preventing overfitting and lowers the network's memory and processing needs. Another glaring benefit discovered through data exploration is that DWT makes it easier to combine data from various frequency sub-bands, enabling the network to learn richer representations of the input data and enhancing feature extraction.

\begin{figure}[htbp]
\centering
\includegraphics[width=0.5\textwidth]{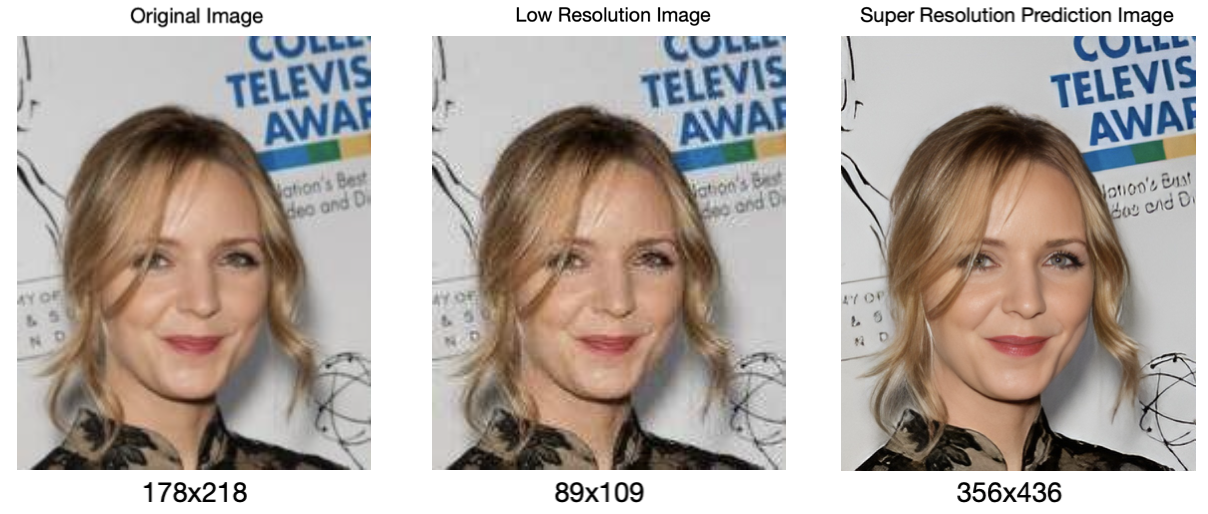}
\caption{Sample output on the CelebA dataset}
\label{fig9}
\end{figure}


\subsection{Results on the CelebA Dataset}
By employing the L-Waveblock on the benchmark CelebA dataset, we obtain the PSNR value of 29.45, against the value of 27.309 without using the L-Waveblock, as seen in Table II. The SSIM value too, is better using the L-WaveBlock than the one without employing it. These metrics quantitatively demonstrate the efficiency of the method in improving the image resolution. The use of DWT in feature extraction facilitates image resolution since DWT decomposes an image into a set of wavelet coefficients at different scales or resolutions. It does this by splitting the image into low-frequency and high-frequency components. The low-frequency component represents the coarse details of the image, while the high-frequency components capture finer details.  DWT can help in extracting important features from an image by analyzing its different frequency components, thereby increasing the image quality and resolution. 

\begin{figure}[htbp]
\centering
\includegraphics[width=0.5\textwidth]{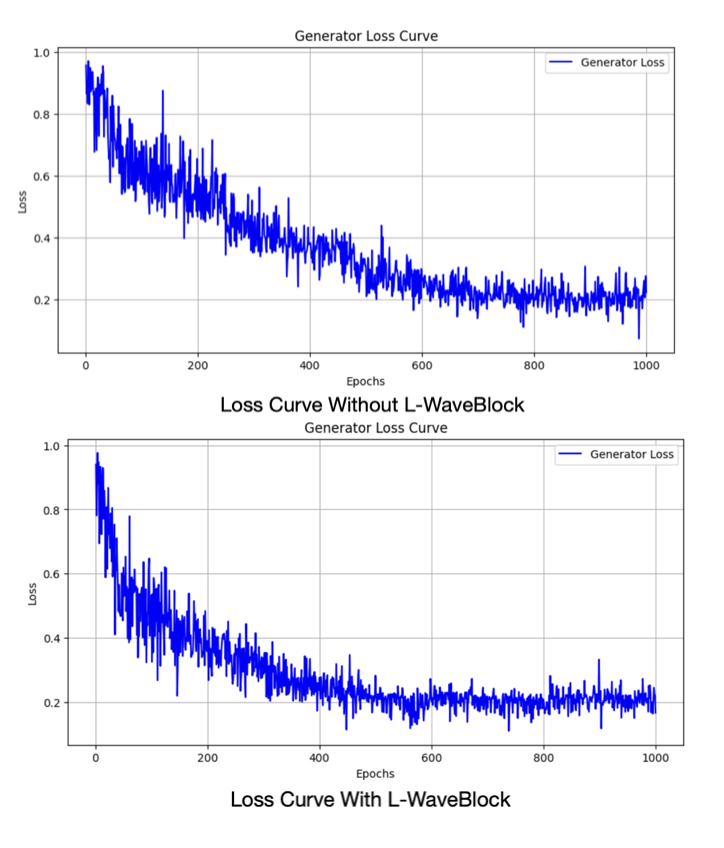}
\caption{Loss curve with and without L-WaveBlock for SRGAN dataset}
\label{fig10}
\end{figure}

\begin{table}
\centering
\caption{Performance metrics for SR-GAN}
\begin{tabular}{lcc}
\toprule
\textbf{Architectures} & \textbf{PSNR /dB} & \textbf{SSIM} \\
\midrule
\textbf{Generator without L-WaveBlock} & 27.309 & 0.853 \\
\textbf{Generator with L-WaveBlock} & \textbf{29.05} & \textbf{0.874} \\
\bottomrule
\end{tabular}
\end{table}

It retains the low-frequency components in both the horizontal and vertical directions. LL sub-band usually captures the overall structure and main features of the image, providing a low-resolution representation. The HL sub-band contains high-frequency information in the horizontal direction and low-frequency information in the vertical direction. It represents details that vary horizontally but are relatively smooth vertically. Features like edges or textures that are primarily horizontal are often found in this sub-band. The LH sub-band contains low-frequency information in the horizontal direction and high-frequency information in the vertical direction. It represents details that vary vertically but are relatively smooth horizontally. This can be seen from Fig 9, where the super-resolution predicted image contains enhanced features than those obtained from the low-resolution or original images. Fig 10 demonstrates the loss convergence for the proposed methodology on the CelebA dataset. 

\begin{table}
\centering
\caption{Performance metrics for denoising on the GoPro dataset}
\begin{tabular}{lcc}
\toprule
\textbf{Architectures} & \textbf{PSNR /dB} & \textbf{SSIM} \\
\midrule
\textbf{Generator without L-WaveBlock} & 26.87 & 0.776 \\
\textbf{Generator with L-WaveBlock} & \textbf{26.913} & \textbf{0.782} \\
\bottomrule
\end{tabular}
\end{table}

\subsection{Results on the GoPro Dataset}
Fig 11 shows the results obtained on the GoPro dataset by utilizing the proposed methodology. The input image is a Gaussian blurred image and by using the DWT and the sub-bands present in its coefficients, we obtain the output which is very closely similar to the ground truth. It can also be seen from Table III that with the usage of the L-WaveBlock on the GoPro dataset, we obtain a PSNR of 26.233, as opposed to the value of 26.87 without the L-WaveBlock. The SSIM value for the proposed methodology is 0.712 in contrast to the value of 0.776 without the L-WaveBlock. Although the results aren't theoretically better, they are competent with the state-of-the-art. Fig 12 shows the loss convergence curve for the same

\begin{figure}[htbp]
\centering
\includegraphics[width=0.4\textwidth]{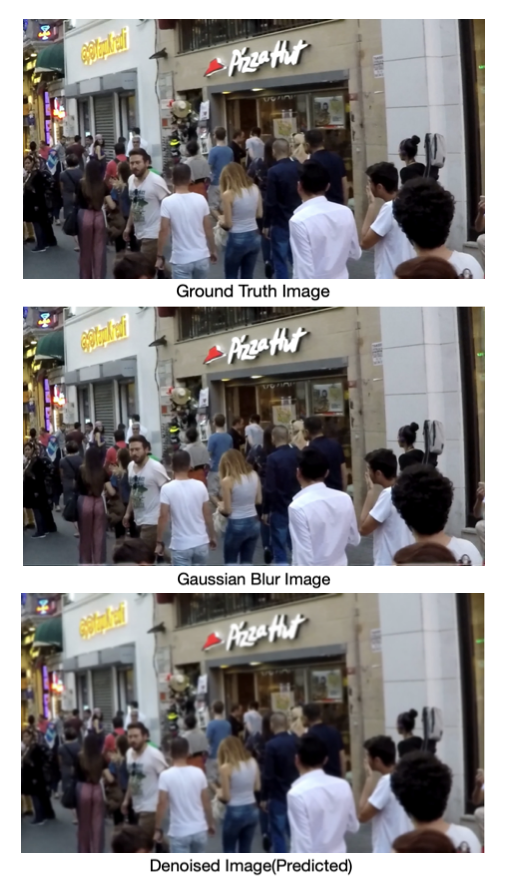}
\caption{Results on the GoPro dataset}
\label{fig11}
\end{figure}

\begin{figure}[htbp]
\centering
\includegraphics[width=0.5\textwidth]{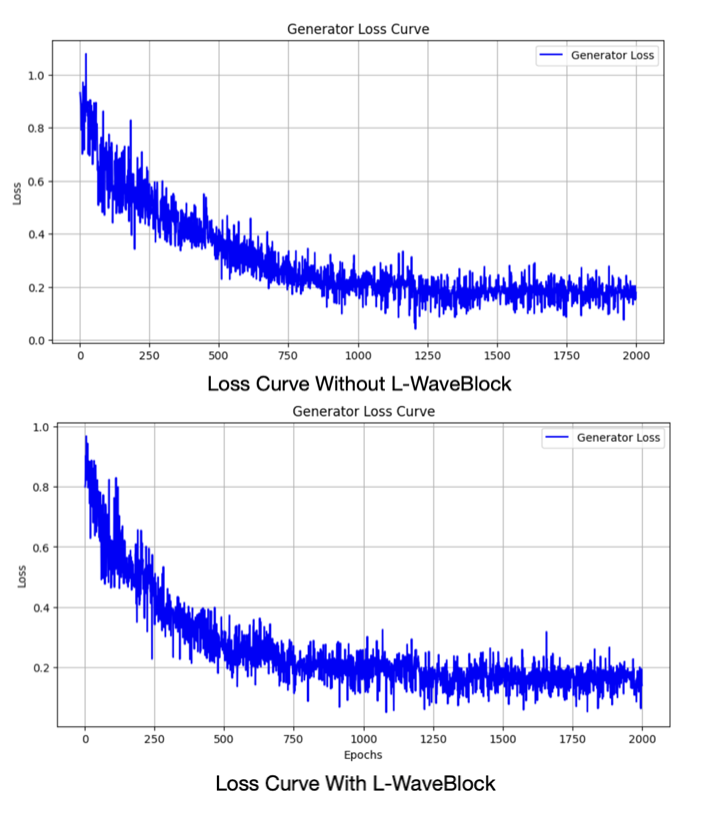}
\caption{Loss curve with and without L-WaveBlock for GoPro dataset}
\label{fig12}
\end{figure}

Additionally, the proposed approach offers two significant advantages in terms of the loss function. Firstly, it exhibits remarkable stability in comparison to other models. This stability ensures that the model trains smoothly without drastic fluctuations or divergences during training. Secondly, the proposed model demonstrates superior learning efficiency, achieving better convergence. While the model reaches optimal performance around 750 epochs, competing models typically require around 1000 epochs to achieve similar results. This accelerated learning pace highlights the effectiveness of our DWT-enhanced GAN in producing high-quality outputs more rapidly.

\section{Conclusion}
In this study, we have introduced a transformative approach that leverages DWT to enhance the feature extraction capabilities of GANs. Across three critical image manipulation tasks - Satellite Image to Maps Image Conversion, Image Denoising, and Image Resolution Enhancement - our method has consistently yielded results that are competent with state-of-the-art outcomes. Notably, its exceptional speed of convergence establishes the superiority of our approach and its usability in practical settings.
Traditional GANs have struggled with longer training times, causing an obstacle to their practicality. Our method, driven by efficient feature extraction through DWT, not only offers a solution to this challenge but also finds a way for a more efficient, accessible, and transformative future in computer vision applications. By placing feature extraction at the forefront and leveraging the multi-resolution analysis capabilities of DWT, our approach suggests a new method in image manipulation where high-quality, contextually meaningful results are achievable with exceptional efficiency.

This research represents a significant stride forward in the realm of computer vision.  Our method ensures faster convergence and has the potential to reshape the landscape of computer vision applications. We look forward to further explorations and applications of this approach, anticipating a future where high-quality, contextually meaningful image manipulation becomes not only achievable but also efficient and accessible to all.

\printbibliography

\end{document}